\pdfoutput=1

\documentclass[11pt]{article}

\usepackage[]{EMNLP2022}

\usepackage{times}
\usepackage{latexsym}
\usepackage{soul}
\usepackage{rotating,tabularx}
\usepackage{url}
\usepackage{amsmath}
\usepackage{booktabs}
\usepackage{bm}
\usepackage{multirow}
\usepackage{subfigure}
\usepackage{algorithm}  
\usepackage{CJK}
\usepackage{algpseudocode}  
\usepackage{makecell}
\usepackage{xspace}
\usepackage{amssymb}
\usepackage{array}
\usepackage{footmisc}

\usepackage{listings}
\usepackage{paralist}
\usepackage{enumitem}
\usepackage{colortbl}
\usepackage{soul}
\usepackage{epstopdf}
\usepackage{setspace}
\usepackage{pifont}
\usepackage{xcolor}
\usepackage{mathrsfs}
\usepackage{todonotes}
\usepackage{newfloat}
\usepackage{listings}
\usepackage[T1]{fontenc}

\usepackage[utf8]{inputenc}

\usepackage{microtype}

\usepackage{inconsolata}

\newcommand{\dubbelop}{$^{\blacktriangle}$}
\newcommand{\ie}{\emph{i.e.,}\xspace}
\newcommand{\hlc}[2][yellow]{{%
		\colorlet{foo}{#1}%
		\sethlcolor{foo}\hl{#2}}%
}
\title{Scientific Paper Extractive Summarization Enhanced by Citation Graphs}

\author{Xiuying Chen$^{1*}$, Mingzhe Li$^{2*}$\thanks{* Equal contribution.}, Shen Gao$^{3}$, Rui Yan$^{4}$,  Xin Gao$^{1\dagger}$,   Xiangliang Zhang$^{5,1\dagger}$\thanks{$\dagger$ Corresponding author.}\\
\\
$^1$\ Computational Bioscience Reseach Center, KAUST\\
$^2$\ Ant Group\\
$^3$\ School of Computer Science and Technology, Shandong University\\
$^4$\ Gaoling School of Artificial Intelligence, Renmin University of China\\
$^5$\ University of Notre Dame\\
\texttt{xiuying.chen@kaust.edu.sa, limingzhe.lmz@antgroup.com}}
\renewcommand\footnotemark{}

\begin{document}
\maketitle
\begin{abstract}
In a citation graph, adjacent paper nodes share related scientific terms and topics. 
The graph thus conveys unique structure information of document-level relatedness that can be utilized in the paper summarization task, for exploring beyond the intra-document information.
In this work, we focus on leveraging citation graphs to improve scientific paper extractive summarization under different settings.
We first propose a \textit{Multi-granularity Unsupervised Summarization} model (MUS) as a simple and low-cost solution to the task.
MUS finetunes a pre-trained encoder model on the citation graph by link prediction tasks.
Then, the abstract sentences are extracted from the corresponding paper considering multi-granularity information.
Preliminary results demonstrate that citation graph is helpful even in a simple unsupervised framework.
Motivated by this, we next propose a \textit{Graph-based Supervised Summarization} model (GSS) to achieve more accurate results on the task when large-scale labeled data are available.
Apart from employing the link prediction as an auxiliary task, GSS introduces a gated sentence encoder and a graph information fusion module to take advantage of the graph information to polish the sentence representation.
Experiments on a public benchmark dataset show that MUS and GSS bring substantial improvements over the prior state-of-the-art model.

\end{abstract}

\section{Introduction}
Text summarization is to automatically glean the most important concepts from an article, removing secondary or redundant concepts.
Among various summarization tasks such as news summarization \cite{wang2020heterogeneous}, dialog summarization \cite{zhang2021unsupervised}, and timeline summarization \cite{chen2019learning}, scientific paper summarization remains a challenging task, since scientific papers are usually longer, and full of complex concepts and domain-specific items in specific fields \cite{cohan2020specter,an2021enhancing}.

\begin{figure}[t]
	\centering
	\includegraphics[width=0.95\linewidth]{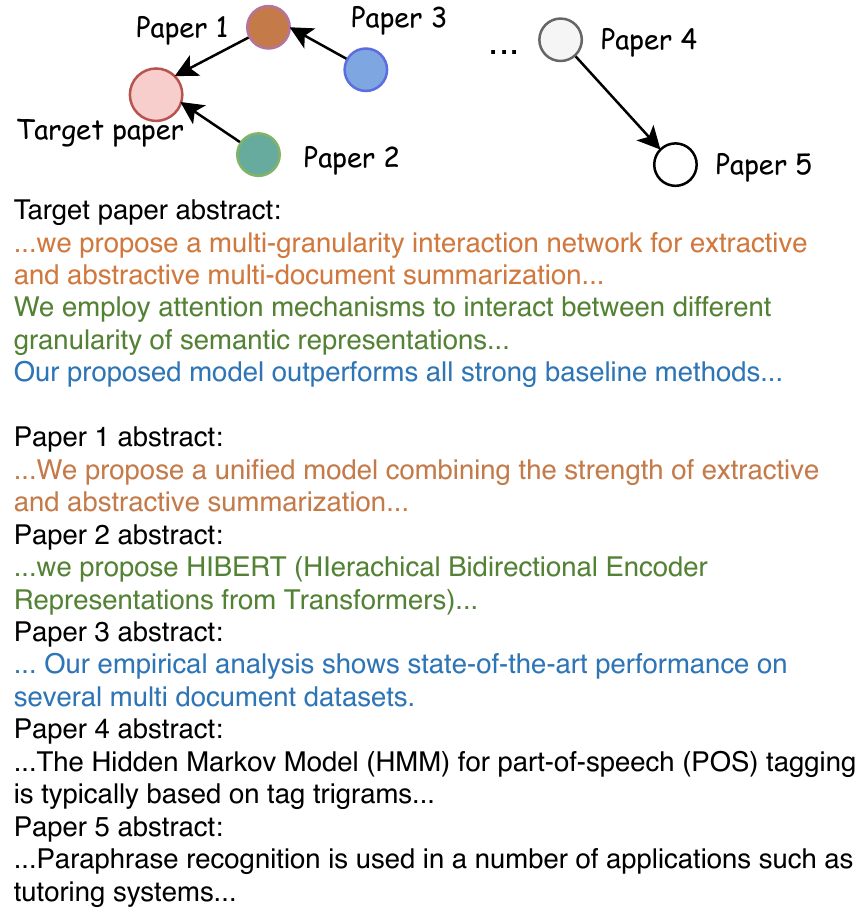}
	\caption{A small research community on the subject of text summarization.
	Arrows denote citation relationships.
	Sentences with shared domain-specific terms are highlighted by the same color (best viewed in color).}
	\label{fig:intro}
\end{figure}

Existing works on extractive scientific paper summarization mainly focus on utilizing intra-document relationships.
For example, 
\citet{xiao2019extractive} divided papers into sections and incorporated the global and local context for extractive summarization.
\citet{dong2021discourse} proposed an unsupervised graph-based ranking model, which assumes a hierarchical graph representation of the scientific papers.
However, the effectiveness of the citation graph in extractive summarization tasks is left to be explored.

In the citation graph domain, nodes are inherently linked and dependent on each other.
Correspondingly, we assume that a good scientific paper abstract should be able to capture this structural information.
Figure~\ref{fig:intro} demonstrates this intuition.
A good paper abstract of the query paper is more relevant to the directly cited paper 1, 2, and indirectly cited paper 3 by sharing domain-specific terms, and not relevant to paper 4 and 5 that are far from it in the citation graph.

Based on this observation, we first propose a \textit{Multi-granularity Unsupervised Summarization} (MUS) as a light solution to the task without the requirement of rich supervision information. 
MUS first finetunes a pre-trained encoder model on the citation graph to obtain better sentence and document representations by solving a link prediction task.
Then, MUS transforms the citation graph into a multi-granularity graph.
Sentences are then extracted considering the multi-granularity position-aware centrality.
Experimental results demonstrate that citation graph information can enhance the summarization even by this simple unsupervised framework.

Motivated by this, we further propose a \textit{Graph-based Supervised Summarization} model (GSS) to achieve more accurate results for the scenario where large-scale labeled data is feasible.
Firstly, a graph neural network encoder based on a pretrained language model is employed to obtain sentence representations for the target document and document representations for related papers in the citation graph.
Secondly, we propose a gated sentence encoder that polishes the sentence representations based on their relatedness to the document gist.
Then, a graph information fusion module is utilized to incorporate the information from reference papers to the polished sentence representations.
Finally, a multi-task framework is applied to the model, which jointly assigns selection weights to extract abstracts and predicts whether there exists an edge between two nodes.
Under the supervision setting, the graph information gives stronger guidance in two ways.
In one way, we employ the link prediction task on the graph to obtain better document representation, and in the other way, the document representation is used to polish sentence representations.
Results show that our model significantly surpasses the prior state-of-the-art model by on public benchmark dataset.

Our contributions can be summarized as follows: 

$\bullet$ Our work demonstrates the effectiveness of citation graph modeling in scientific paper extractive summarization.

$\bullet$ We propose an unsupervised summarization model MUS and a supervised model GSS, both of which learn from the citation graph structure to achieve better summarization.

$\bullet$ We bring new state-of-the-art unsupervised and supervised performance on a public scientific summarization dataset. 
We release our code for further research.

\section{Related Work}

Extractive summarization aims to generate a summary by integrating the salient sentences in the document.
Traditional extractive summarization methods are mostly unsupervised, which directly select sentences based on explicit features or graph-based methods.
Unsupervised methods can save a lot of manual work and expenses since it does not demand labeled data.
More recently, benefiting from the success of neural sequence models and large-scale datasets, neural network-based supervised summarization models have been proposed \cite{gao2019write,gao2020standard}.
More recetly, \citet{liu2019text} and \citet{xu2020discourse} showcased how pre-trained models can be applied in extractive text summarization tasks.
These models often achieve better performance since labeled data is provided.

Research on summarizing scientific articles has been studied for decades \cite{nenkova2011automatic}.
In the unsupervised domain,
\citet{cohan2015scientific}  proposed to extract citation-contexts from citing articles, which does not help draft an abstract when the paper has not been cited yet.
\citet{dong2021discourse} came up with an unsupervised graph-based ranking model for extractive summarization.
As for supervised methods, \citet{subramanian2019extractive} performed an extractive step before generating a summary, which is then used as guidance for abstractive summarization.
\citet{xiao2019extractive} incorporated both the global context and local context to summarize papers.

Early approaches for extractive summarization, such as LexRank and TextRank \cite{mihalcea2004textrank}, have taken advantage of graph structures with inter-sentence cosine similarity.
As for the neural-based approaches,
\citet{koncel2019text} designed a graph attention-based transformer encoder to generate a summary with the help of knowledge graphs extracted from scientific texts.
Graph modeling is also explored in abstractive summarization.
For example, \citet{an2021enhancing} proposed a citation graph-based summarization model which incorporates inter-document information of the source paper and its references.


\section{Problem Formulation}
\label{sec:formulation}

We define a citation graph on the whole dataset, which contains scientific papers and citation relationships. 
Each node represents a scientific paper in the dataset, and each edge indicates the citation relationship between two papers. 
Each paper in the graph consists of a list of sentences $ s = \{s_{1},...,s_{T_s}\}$, where $T_s$ is the number of sentences in the document. 
Our model aims to generate a score vector $\mathbf{y} = \{y_{1},...,y_{T_s}\}$ for each sentence, where each score $y_{i} \in [0,1]$ denotes the sentence's extracting probability.
For the unsupervised model, the model outputs the $\mathbf{y}$ score without supervision.
For the supervised model, we annotate the given sentences by a gold label vector $\mathbf{y'}$.
During the training process, the cross-entropy loss is measured and minimized between $\mathbf{y}$ and  $\mathbf{y'}$.

\section{Multi-granularity Unsupervised Summarization Model}

In this section, we propose our \textit{Multi-granularity Unsupervised Summarization} model (MUS). 
MUS combines a finetuned pre-trained encoder with a multi-granularity model on the citation graph.

\begin{figure}
	\centering
	\includegraphics[width=1\linewidth]{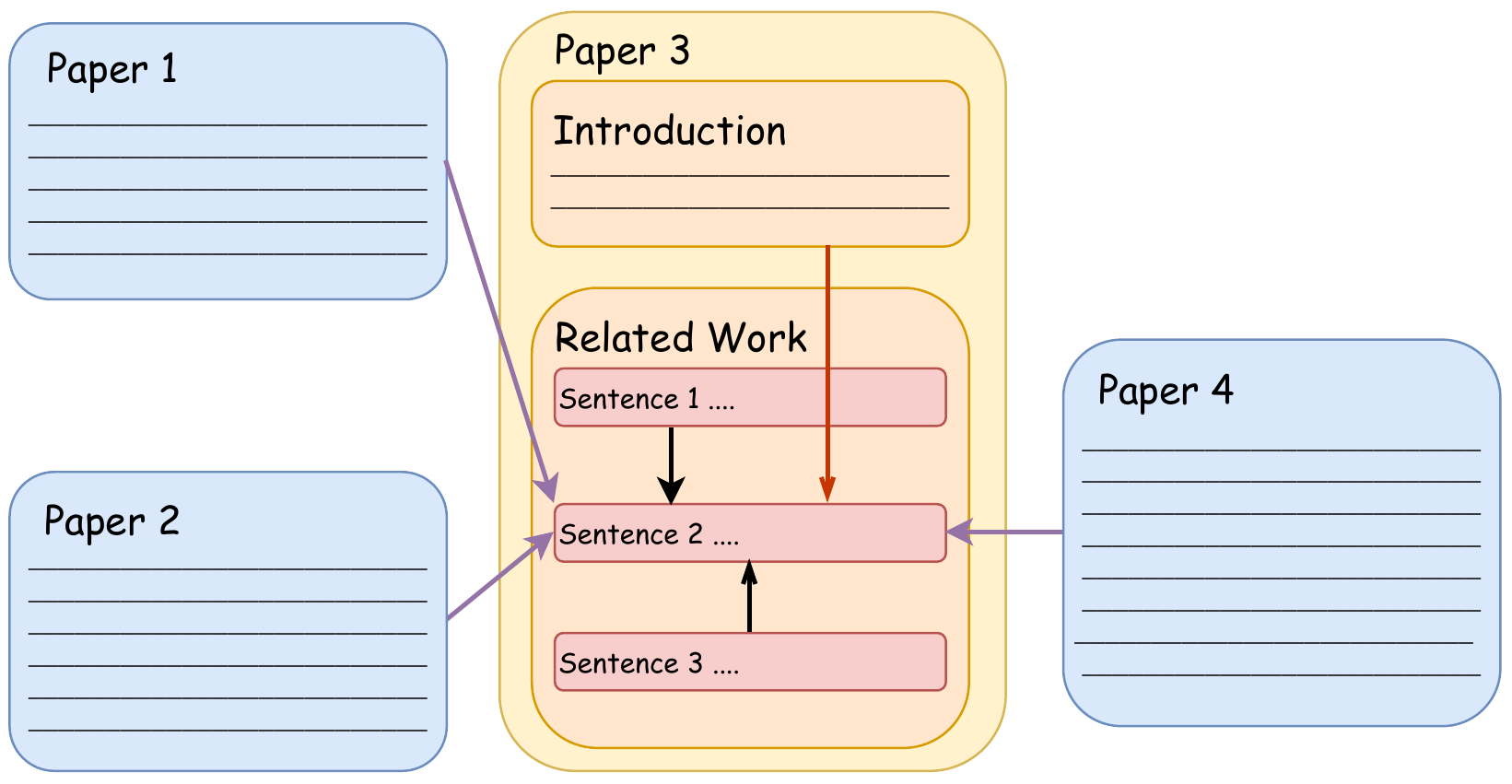}
	\caption{Example of a multi-granularity graph for unsupervised summarization (paper 1,2, and 4 are cited by 3).
    We select sentence 2 in paper 3 as the centered node to denote the multiple granularities (sentence-sentence, sentence-section, and sentence-document). }
	\label{fig:un}
\end{figure}

\subsection{Finetuned SciBERT Encoder}
\label{scibert}

SciBERT~\cite{beltagy2019scibert} is a pre-trained language model which leverages unsupervised pretraining on scientific publications.
However, different from plain text, in the citation graph domain, papers are inherently linked and dependent on each other, which provides rich information that enables us to design pretext tasks such as graph structure prediction \cite{jin2020self}.
Hence, we exploit a type of graph-level distributional hypothesis to finetune SciBERT, which is to predict whether or not there exists a link between a given node pair. 
In particular, each training instance is a triplet of papers: the query paper $\mathbf{d}$, a positive paper $\mathbf{d}^+$ and a negative paper $\mathbf{d}^-$. 
The positive paper is a paper that the query paper cites, and the negative paper is a paper that is not cited. 
The query paper representation $\mathbf{d}$ is taken as the average of sentence representations in the document, and other paper representations $\mathbf{d}^+$ and $\mathbf{d}^-$ are the averages of sentences in the abstract.
We train the model using contrastive learning:
\par
{\small
\begin{align}
    \mathcal{L}&=-\log \operatorname{sim}(\mathbf{d},\mathbf{d}^+)-\log[1-\operatorname{sim}(\mathbf{d},\mathbf{d}^{-})],\label{cross}
\end{align}
}%
where $\operatorname{sim}$ is the cosine distance function. 
The objective is to distinguish adjacent papers from other papers in the corpus, and the encoder is pushed to capture the meaning of the intended paper in order to achieve that. 
If the document representations can be used to correctly predict a link, then these representations are of good quality, and so are the sentence representations that form the document representations.

\subsection{Multi-granularity Graph }
MUS is a graph-based summarization framework \cite{mihalcea2004textrank}, where sentence nodes with higher centrality are more likely selected in the abstract.
We not only consider intra-document sentence centrality in the graph, but also consider inter-document sentence centrality (see Figure~\ref{fig:un}).


First, to model the local importance of a sentence within its section, we create a fully-connected subgraph for each section, where we allow sentence-sentence edges for all sentences. 
The importance of a sentence $s_i$ is determined by the sum of edge weights connecting $s_i$ to other sentences: $    a_i^{\text{sen}}=\textstyle \sum_{j=1}^{n_{\text{sen}}} e^{\text{sen}}_{ij}$.
Each edge weight $e^{\text{sen}}_{ij}$ is defined by:
\begin{align}
    e_{ij}^{\text{sen}}=\left\{\begin{array}{ll}\lambda_{1} * \operatorname{sim}(\mathbf{s}_i, \mathbf{s}_j), & \text { if } b_i<b_j \\ \lambda_{2} * \operatorname{sim}(\mathbf{s}_i, \mathbf{s}_j), & \text { if } b_i\geq b_j\end{array}\right.
\end{align}
where $b_i$  measures the boundary position of $s_i$ among $n_{\text{sen}}$ sentences in  the section: $ b_i=\min (\text{loc}_i, n_{\text{sen}}-\text{loc}_i)$.
$\text{loc}_i$ is the location index of  $s_i$ in this section. 
The closer the sentence is to the section boundary, the lower boundary score it will have. The definition of $e_{ij}^{\text{sen}}$,
with hyperparameters $\lambda_{1}<\lambda_{2}$, reflects two principles on measuring sentence local importance. Firstly, a sentence that is similar to a greater number of other sentences in the same section should be more important \cite{mihalcea2004textrank}, incorporated by measuring $\operatorname{sim}(\mathbf{s}_i, \mathbf{s}_j)$ on two sentence embeddings. Secondly, sentences with crucial information are more likely to appear in the start and end sentences of a text span \cite{Lin1997IdentifyingTB,Teufel1997SentenceEA}, operated by applying a larger $\lambda_{2}$ when $b_j$ is small ($b_i\geq b_j$).

Next, we measure the global importance of a sentence $s_i$ with respect to $n_{\text{sec}}$ sections in the document based on the weights of section-sentence edges:
\begin{align}
 & a_i^{\text{sec}}=\textstyle \sum_{r=1}^{n_{\text{sec}}}    e^{\text{sec}}_{ir},   \\
 & e^{\text{sec}}_{ir}=\left\{\begin{array}{ll}\lambda_{1} * \operatorname{sim}\left(\mathbf{s}_i,\mathbf{o}_r\right),  \text { if } \hat{b}_k<\hat{b}_r   \\ 
  \lambda_{2} * \operatorname{sim}\left(\mathbf{s}_i,\mathbf{o}_r\right),  \text { if } \hat{b}_k \geq \hat{b}_r\end{array}\right. 
\end{align}
where $\mathbf{o}_r$ is the representation of $r$-th section node, which is initialized by the average pooling of sentence representations.
We use $k$ to denote the index of the section sentence $s_i$ belongs to.
$\hat{b}$ is calculated similarly to $b$, but on section level.
A section with a lower $\hat{b}$ score is closer to the boundary of the document and will have a higher edge weight.
Note that we only allow section-sentence edges instead of sentence-sentence across different sections for efficiency sake.


Last, we model the interaction between documents by the distance between the sentences in the target document $d$ and the reference document $d^+$:
\begin{align}
  a_i^{\text{doc}} = \textstyle \sum_{d^+} e^{\text{doc}}_{i+}= \textstyle \sum_{d^+} \operatorname{sim}(\mathbf{s}_i,\mathbf{d}^+).
\end{align}

\subsection{Importance Calculation}
The overall importance of a sentence $s_i$ is computed as the weighted sum of its multi-granularity centrality scores: $c_i= \mu_1 a_i^{\text{sen}} + \mu_2 a_i^{\text{sec}} + \mu_3 a_i^{\text{doc}}$, 
where 
$\mu_{1}, \mu_2,\mu_3$ are weighting factors set for centralities of different granularity.

\section{Graph-based Supervised Summarization Model}

\begin{figure}
	\centering
	\includegraphics[width=0.75\linewidth]{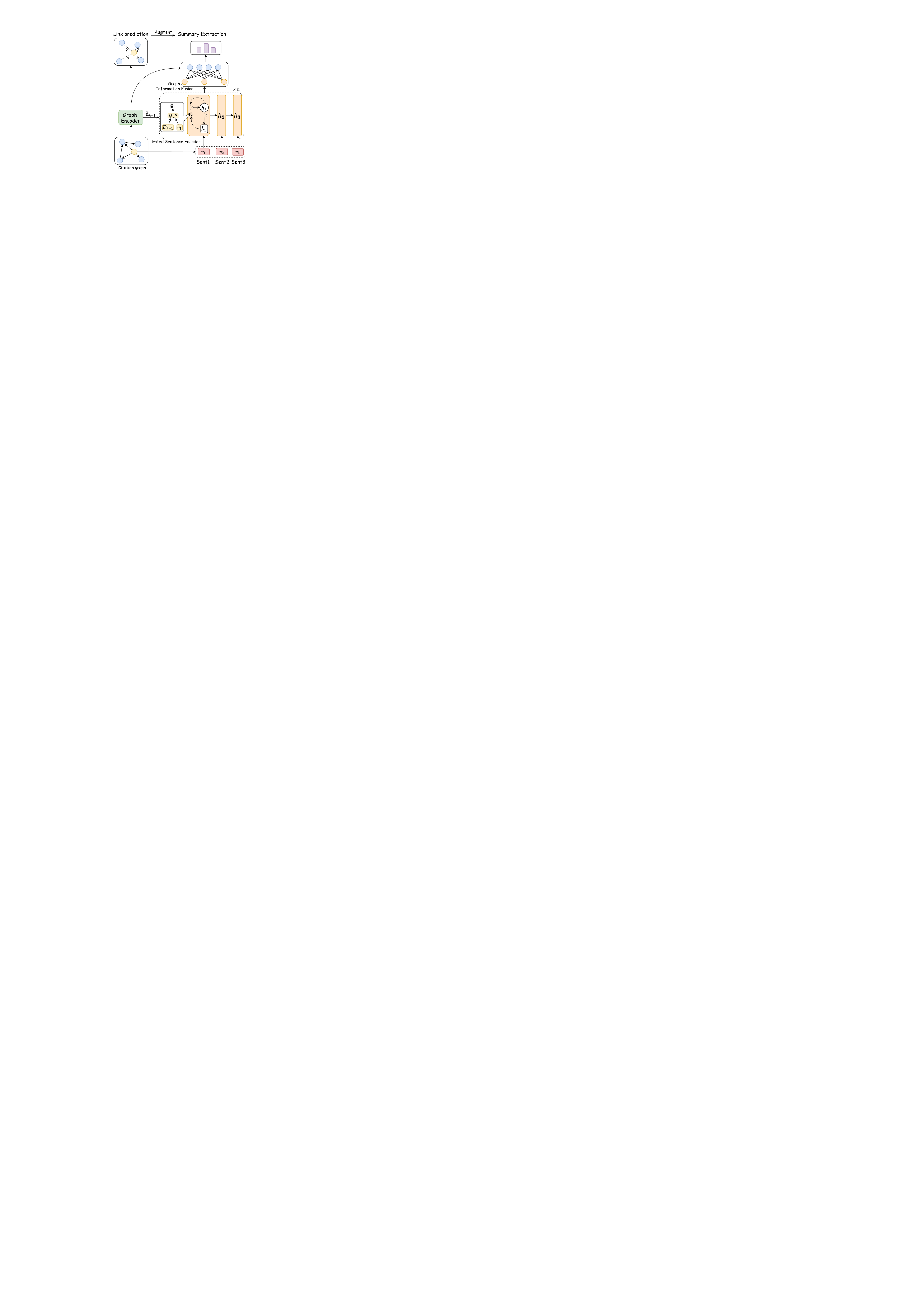}
	\caption{Graph-based supervised summarization model.}
	\label{fig:su}
\end{figure}

We then introduce a \textit{Graph-based Supervised Summarization} model (GSS), which uses the graph structure to extract abstracts under supervision, as shown in Figure \ref{fig:su}.

\subsection{Graph Encoder}
We propose a graph encoder that is built on the pre-trained SciBERT encoder.
We first use SciBERT to obtain sentence representation $\mathbf{s}_i$ and document representation $\mathbf{d}^0$ similar to MUS.
Then, we employ GraphSAGE \cite{hamilton2017inductive} to update the contextual representation of each paper:
\par
{\small
\begin{align}
    \mathbf{d}^{l+1}&=\operatorname{aggregate}\left(\left\{\mathbf{d}^l_{j}, \forall d_j \in \mathcal{N}(d)\right\}\right),\\ 
    \mathbf{\tilde{d}}^{l+1}&=\sigma\left(W_{\mathbf{v}} \cdot [\mathbf{d}^{l}, \mathbf{d}^{l+1}]\right),
    \mathbf{\hat{d}}^{l+1} =\mathbf{\tilde{d}}^{l+1}/||\mathbf{\tilde{d}}^{l+1}||_2,
\end{align}
}%
where $\mathcal{N}(d)$ is the reference documents of the target paper, $[,]$ denotes the concatenation operation, $l$ is the layer index, and $\operatorname {aggregate}$ is the differentiable aggregator function.
For brevity, we denote the final output for the document as $\mathbf{\hat{d}}$.

\subsection{Gated Sentence Encoder}
Previous works show that document-level information can be useful for extractive summarization \cite{wang2020heterogeneous,jin2020multi}.
Hence, we propose a gated sentence encoder, which polishes the sentence representation $\mathbf{s}_i$ in the target paper under the guidance of the target document $\mathbf{\hat{d}}$.

Our gated encoder is based on a modified version of GRU that takes sentences as inputs.
The update gate vector $\mathbf{z}_i$ in GRU only takes the sentence representation as input, without considering document level information. 
To this end, we propose a new reading gate $\mathbf{g}_i$. 
The calculation of $\mathbf{g}_i$ takes in two inputs, the sentence representation and the document representation, to highlight the input texts which heavily correlate with the document gist:
\begin{align}
   \mathbf{f}_i&=\left[\mathbf{s}_i \circ \mathbf{h}_i; \mathbf{s}_i; \mathbf{\hat{d}}\right],\\
   \mathbf{F}_i&=W_{b} \tanh \left(W_{\mathbf{a}} \mathbf{f}_i+b_{\mathbf{a}}\right)+b_{\mathbf{b}},\\
   \mathbf{g}_i&=\textstyle\frac{\exp \left(\mathbf{F}_i\right)}{\sum_{k=1}^{T_s} \exp \left(\mathbf{F}_k\right)},\\
   \mathbf{h}_{i+1}&=\mathbf{g}_i \odot \mathbf{\hat{h}}_i+\left(1-\mathbf{g}_i\right) \odot \mathbf{h}_i,\label{sru}
\end{align}
{where $T_s$ is number of sentences in the document.}

This gated reading process is a multi-hop process and we use superscript $k$ to denote the hop index.
After each sentence passes through the selective reading module, we wish to update the document representation $\mathbf{\hat{d}}^{k}$ with the newly constructed sentence representations.
We use a $\text{GRU}_{\text{upd}}$ cell to generate the polished document representation, whose input is the final state of the gated network from the previous hop: $\mathbf{\hat{d}}^{k+1}=\text{GRU}_{\text{upd}}(\mathbf{h}^{k}_{T_s},\mathbf{\hat{d}}^{k})$.

The final output of sentences is obtained with an additional residual layer $\mathbf{s}_i^{\text{final}}=\mathbf{h}_i^{K}+\mathbf{s}_i$.

\subsection{Graph Information Fusion}
In this subsection, we update the sentence representations by paying attention to its cited documents.
The intuition is that, in practice, researchers usually write an abstract of a paper by referring to the related papers. 
We thus apply a graph information fusion module, which performs Multi-Head Attention (MHA) \cite{vaswani2017attention} across the polished sentence representations $\mathbf{s}^{\text{final}}_i$ and $m$ randomly sampled cited paper representations $\mathbf{\hat{d}}^k$ to obtain the global sentence representation $\mathbf{\hat{a}}_i$, where the query is $\mathbf{s}^{\text{final}}_i$, and key and value are $\mathbf{\hat{d}}^k$.

\subsection{Link Prediction}

In a supervised learning setting, the intuition that good sentence representation can help link prediction still holds.
What is more, good document representations can help polish the sentence representations as introduced in the gated sentence encoder.
Hence, we employ link prediction as an auxiliary task, which predicts whether there is an edge between the $i$-th and the $j$-th document based on representations from the graph encoder module.
Details are similar to the setting in MUS, with loss function defined in Equation \ref{cross}.

Finally, we build a classifier to select sentences based on the sentence representations $\mathbf{\hat{a}}_i$: ${y_i}=\sigma\left(\mathbf{\hat{a}}_i W_{\mathbf{o}}+b_{\mathbf{o}}\right)$.

\section{Experimental Setup}
\label{sec:experiment}

\subsection{Dataset}
We evaluate our model on a Semantic Scholar Network dataset (SSN), proposed by \citet{an2021enhancing}.
SSN has two versions, \ie SSN (inductive) and SSN (transductive).
For SNN (transductive),  6,250/6,250 papers are randomly chosen from the whole dataset as test/validation sets, and the remaining 128,299 papers are employed as the training set.
On the contrary, SNN (inductive) splits the whole citation graph into three independent subgraphs, considering that in real cases, the test papers may form a new graph that has no overlap with the training dataset.
The training/validation/test graphs in the inductive setting contain 128,400/6,123/6,276 nodes and 603,737/17,221/14,515 edges, respectively. 

\subsection{Comparison Methods}

To evaluate the performance of our proposed models,  we compare them with the following unsupervised summarization and supervised summarization baselines.

\noindent \textbf{\textit{Unsupervised summarization baselines}}:

\noindent
(1) \texttt{LEAD}: extracts the first $L$ (depending on the number of sentences in the reference abstract) sentences from the source document.
(2) \texttt{PACSUM}~\cite{zheng2019sentence}: revisits the popular graph-based ranking algorithm and modifies how node (sentence) centrality is computed.
(3)
\texttt{HipoRank}~\cite{dong2020hiporank}: leverages positional and hierarchical information grounded in the discourse structure to augment a document’s graph representation.

\noindent \textbf{\textit{Supervised summarization baselines}}:

\noindent
 (1) \texttt{BertSumEXT}~\cite{liu2019text}: the extractive summarization model with BERT as the encoder.
(2) \texttt{MGSum-\textit{ext}}~\cite{jin2020multi}: the extractive multi-document summarization baseline, which extracts sentences from both the document and the abstracts of the reference papers.
We incorporate this baseline to see if sentences from the reference papers are useful.
 (3) \texttt{EMSum}~\cite{zhou2021entity}: the abstractive multi-document summarization model, which takes the paper with references as multiple input documents.
(4) \texttt{CGSum}~\cite{an2021enhancing}: the abstractive summarization baseline with citation graph as input, which is also the paper that proposes the SSN dataest.
(5) \texttt{HSG}~\cite{wang2020heterogeneous}: the heterogeneous graph-based neural network for extractive summarization.

\begin{table*}[t!]
		\small
		\centering
		\resizebox{140mm}{!}{\begin{tabular}{l ccccc | ccccc }
			\toprule
			\multirow{3}{*}{Models} & \multicolumn{5}{c}{\textsf{ SSN (inductive)}} & \multicolumn{5}{c}{\textsf{SSN (transductive)}} \\
			\cmidrule(lr){2-6}  \cmidrule(lr){7-11} 
			& RG-1  & RG-2  & RG-L &BERTSCORE &QuestEval & RG-1  & RG-2  & RG-L &BERTSCORE &QuestEval\\
			\midrule
			\textit{oracle ext}      &  51.04  &23.34 &45.88  & - & - &  50.12  &  23.31& 45.29  & - & -\\
			\midrule
			\multicolumn{4}{@{}l}{\emph{Unsupervised methods}}\\
			LEAD      &  28.29  &5.99 & 24.84 &  82.22 & 31.74 &  28.30 &6.87 & 24.93 & 82.11 & 30.52 \\
			TextRank    &  36.36&9.67 &  32.72 &  83.53 & 31.94 &  40.81& 12.81& 36.47 & 83.75& 32.07 \\
			
			PACSUM   &  37.74&9.52&
            34.74 &  83.70 & 32.01 & 40.83 & 12.16 & 36.81 & 83.93& 32.13  \\
			
			HipoRank     & 40.37& 11.98& 36.58 & 84.27 & 32.23 & 40.56& 12.36& 36.83 & 84.07 & 32.58\\
			
			{\bf MUS} &  \textbf{43.89} & \textbf{13.07} & \textbf{39.04} & \textbf{86.26 }& \textbf{34.19} &  \textbf{42.93} & \textbf{12.85} & \textbf{38.76} & \textbf{85.13} & \textbf{33.90}\\
			\midrule
					
         	\multicolumn{4}{@{}l}{\emph{Supervised methods}}\\

    		BertSumEXT&   44.28 & 14.67&39.77 & 86.58 & 34.24 & 43.23& 14.59 & 38.91 & 85.33 & 34.05\\
    		
    		MGSum-\textit{ext}   &   45.49 &14.87& 40.22 & 86.91 & 34.66 &  43.30  & 14.57 & 39.24 & 85.55 &34.23    \\
    		
    		EMSum   &  44.06 & 14.54 & 39.49 & 86.49 & 34.21 &   43.03 & 14.43 & 38.74 & 85.39 &  34.09   \\
    		
    		CGSum   &   44.28 &14.75& 39.76 &  86.50 & 34.59 & 43.45 & 14.71 & 38.89 &  85.86  & 34.51  \\
    		
    	    HSG & 45.46 & 15.08 &  40.62  & 86.71 & 34.90 & 44.84 &15.35 &  39.72 & 86.04 & 34.62 \\

    	{\bf GSS} & \textbf{47.71} & \textbf{16.78} & \textbf{42.04} & \textbf{88.90} & \textbf{35.68} & \textbf{46.52} & \textbf{16.57} & \textbf{41.85} & \textbf{87.60}  & \textbf{35.64}\\
			
			\midrule
			 \multicolumn{4}{@{}l}{\emph{MUS ablation models}}\\
        w/o finetune  &   43.51 & 12.89 & 38.59 & 85.82  & 33.78 &  42.43 & 12.35 & 38.58 & 84.59 & 33.50\\ 
        w/o position   & 43.61& 13.04 & 38.83 & 85.89 & 33.82 &  42.62 &  12.64 & 38.74 & 84.66 & 33.78 \\
        w/o sentence level  &  43.08 & 12.74 & 38.36 &  85.53 & 33.44 & 42.29 & 12.58& 38.32 & 84.42 & 33.46  \\
        w/o section level  & 43.31& 12.82 & 38.87 & 85.62 &  33.68&  42.58 & 12.66 & 38.56 & 84.62 & 33.71\\ 
        w/o document level & 42.53 & 12.11 & 37.55& 85.17 &33.05 & 41.76  & 11.80 & 37.06 & 84.34 & 33.08 \\ 
        \midrule

        \multicolumn{4}{@{}l}{\emph{GSS ablation models}}\\
        w/o encoder & 46.75 & 16.21 &41.17 & 87.20 & 35.08 & 45.57  & 15.97 &  40.73 & 86.86 & 35.09 \\   
        w/o gated & 46.92 & 16.13 &41.66 & 87.85 & 35.23 & 45.86 & 15.61 & 40.75 & 86.92& 35.36  \\ 
        w/o fusion & 47.23 & 16.24 &41.78 &  88.49 & 35.53 &  46.17 & 16.27 &  41.57 & 87.44 & 35.33 \\ 
        w/o multi  &  47.06 &16.17& 41.44 & 88.22 & 35.47 &  46.08  & 16.33 & 41.49 & 87.26 & 35.28  \\   
			\bottomrule
		\end{tabular}}
		\caption{ROUGE scores comparison between our models, ablation models, and baselines. 
			All our ROUGE scores have a 95\% confidence interval of at most $\pm$0.23.}
			\vspace{-3mm}
		\label{tab:comp_ROUGE_baselines}
	\end{table*}

\subsection{Implementation Details}

For both supervised and unsupervised settings, we implement our experiments in Pytorch on an NVIDIA V100 GPU. 
Following \citet{an2021enhancing}, for all models and baselines, we truncate the input document to 500 tokens. 
For the link prediction task, we sample 5 negative samples for each positive example to approximate the expectation.
For unsupervised models, the batch size is set to 2 (the maximum that fits in our GPU memory).
$\lambda_1$ is set to $0.9$, $\lambda_2$ is $1.0$, 
$\mu_1, \mu_2, \mu_3$ are set to 0.4, 0.1, 0.5, respectively, based on the performance on validation dataset.
For the supervised models, we use batch size 16 and limit the input ground truth abstract length to 150 tokens.
The sample number in the graph information fusion module is also set to 5.
The iteration number in the gated sentence encoder is 3, and the aggregate function is set to the mean operator.
We use Adam optimizer~\cite{kingma2014adam}.
We select the best checkpoint based on the RG-L score on the validation set.

\section{Experimental Results}

\subsection{Overall Performance}

\textbf{Automatic evaluation.}
we evaluate summarization quality by standard ROUGE-1, ROUGE-2, and ROUGE-L ~\cite{lin2004rouge} on full-length F1.
We then use BERTSCORE \cite{zhang2019bertscore} to calculate a similarity score between the summaries based on their BERT embeddings.
We also evaluate with the latest factual consistency metrics, QuestEval \cite{scialom2021questeval}, which considers not only factual information in the generated summary, but also the most important information from the reference.

Table \ref{tab:comp_ROUGE_baselines} shows the results on SSN (inductive) and SSN (transductive).
As can be seen, in the first block, among unsupervised baselines, 
hierarchical and directional graph-based baselines \texttt{PACSUM} and \texttt{HipoRank} achieve better performance than the classic undirectional baseline \texttt{TextRank}, indicating that more comprehensive structure information is stored in the plain text.
Our model MUS achieves significantly better performance compared with the two latest baselines by large margins, which verifies our basic assumption that citation graph is helpful in scientific paper summarization.

The second block in Table \ref{tab:comp_ROUGE_baselines} reports results of the latest supervised baselines and our GSS.
Generally, extractive methods obtain better performance than abstractive baselines.
This demonstrates that scientific corpus tends to use existing words and phrases to form the abstracts instead of using new phrases.
Secondly, it can be seen that pretrained models (\texttt{BertSumEXT}) and graph structure inside the document (\texttt{HSG}) can bring improvements to the vanilla extraction framework.
We also find that incorporating sentences from reference papers cannot increase the oracle score and model performance (\texttt{MGSum}).
This is also intuitively understandable, since sentences from other papers may not be matchable to the current paper.
Finally, our GSS model can make new state-of-the-art performance with the help of citation graph information.

\begin{table}[tb]
	\small
	\centering{\begin{tabular}{lcccc}
			\toprule
			Model& QA(\%) & Info & Coh & Succ \\
			\midrule
			\multicolumn{5}{@{}l}{\emph{Unsupervised models methods}}\\
			PACSUM & 30.8\phantom{0} & 2.12\phantom{0} & 2.14\phantom{0} & 2.18\phantom{0}\\
			 HipoRank & 36.2\phantom{0} & 2.20\phantom{0}  & 2.25\phantom{0} & 2.24\phantom{0}\\
			MUS & 41.4\dubbelop & 2.54\dubbelop & 2.47\dubbelop & 2.32\dubbelop\\
			\midrule
			\multicolumn{5}{@{}l}{\emph{Supervised models methods}}\\
			MGSum-\textit{ext} & 32.9\phantom{0} & 2.18\phantom{0} & 2.15\phantom{0} & 2.20\phantom{0}\\
			 HSG & 38.3\phantom{0} & 2.37\phantom{0}  & 2.40\phantom{0} & 2.33\phantom{0}\\
			GSS & 46.8\dubbelop & 2.62\dubbelop & 2.51\dubbelop & 2.42\dubbelop\\
			\bottomrule
	\end{tabular}}
	\caption{Comparison of human evaluation in terms of QA task, informativeness (Info), coherence (Coh) and succinctness (Succ).
	\dubbelop denotes the improvement to the best baseline is statistically significant (t-test with p-value $\textless$ 0.01).}
	\label{tab:human_evaluation}
\end{table}

    \textbf{Human evaluation.}
    We also assessed the generated results by eliciting human judgments on 40 randomly selected test instances. 
    Our first evaluation quantified the degree to which summarization models can retain the key information following a question-answering paradigm \cite{chen2021capturing}.
	We created 94 questions based on the gold abstract and examined whether participants were able to answer these questions by reading generated abstracts. 
	The questions are written and chosen by two PhD students together.
	Our second evaluation assessed the overall quality of the abstracts by asking participants to score them by taking into account the following criteria: \textit{Informativeness} (Info), \textit{Coherence} (Coh), and \textit{Succinctness} (Succ). 
	Both evaluations were conducted by another three PhD students independently.
	The rating score ranges from 1 to 3, with 3 being the best, and a model’s score is the average of all scores.
	Participants evaluated abstracts produced by baselines that achieved better performance in automatic evaluations.

    As shown in Table \ref{tab:human_evaluation}, participants overwhelmingly prefer our model against comparison systems across datasets.
     The kappa statistics are 0.42, 0.51, and 0.43 for Info, Coh, and Succ, respectively, indicating the moderate agreement between annotators.
    We give a selected case in Table \ref{tab:case}, where our generated abstract captures most key concepts in the gold abstract and summarizes the contribution of the paper in logic.

\begin{table*}[tb]
		\scriptsize
		\begin{center}
			\begin{tabular}{lp{4cm}}
				\toprule  
			\begin{tabularx}{14.2cm}[]{@{}X@{}}
				\textbf{\textit{Ref 1}}: 	we highlight the capabilities of our method with a series of complex liquid simulations , and with a set of single - phase buoyancy simulations . 
				with a set of trained networks , our method is more than two orders of magnitudes \hlc[pink!80]{faster than} a traditional pressure solver .
    			\end{tabularx}
				\\ 	\begin{tabularx}{14.2cm}[]{@{}X@{}}
				\textbf{\textit{Ref 2}}: we introduce the exponential linear unit  ( elu ) which \hlc[cyan!30]{speeds up learning in deep neural networks} and leads to higher classification accuracies . 
				like rectified linear units ( relus ) , leaky relus ( lrelus ) and parametrized relus ( prelus ) , elus alleviate the vanishing gradient problem via the identity for positive values ... 
    			\end{tabularx}
				\\ \begin{tabularx}{14.2cm}[]{@{}X@{}}
				\textbf{\textit{Ref 3}}: ...a \hlc[cyan!30]{convolutional neural network} is trained on a collection of discrete , parameterizable fluid simulation velocity fields ... 
    				reconstructed velocity fields are generated up to \hlc[pink!80]{700x faster} than re - simulating the data with the underlying cpu solver , while achieving compression rates of up to 1300x ."
    			\end{tabularx}
				\\   				\begin{tabularx}{14.2cm}[c]{@{}X@{}}
				\textbf{\textit{Ref 4}}:... an efficient method for the calculation of ferromagnetic resonant modes of \hlc[yellow!40]{magnetic structures} is presented . 
					finite - element discretization allows flexible geometries and location dependent material parameters . 
    			\end{tabularx}
    			\\
    			\hline
    			\begin{tabularx}{14.2cm}[c]{@{}X@{}}
				\textbf{\textit{Gold}}: 
				deep neural networks are used to model the magnetization dynamics in magnetic thin film elements .
						\hlc[yellow!40]{the magnetic states of a thin film element} can be represented in a low dimensional space .
						with \hlc[cyan!30]{convolutional autoencoders} a compression ratio of 1024:1 was achieved . 
						time integration can be performed in the latent space with a second network which was trained by solutions of the landau - lifshitz - gilbert equation . 
						thus the magnetic response to an external field can be computed \hlc[pink!80]{quickly} ."
    			\end{tabularx}
    			\\
    			\hline
    			\begin{tabularx}{14.2cm}[c]{@{}X@{}}
			\textbf{\textit{HSG}}: 
			machine learning has been successfully used in fluid dynamics in order to speed up simulations . 
in this letter we propose a \hlc[cyan!30]{convolutional neural network} to reduce the dimensionality of thin \hlc[yellow!40]{film magnetization} and show how latent space dynamics can be applied to predict the magnetic response of magnetic thin film elements .
finally , we decode the compressed states along the trajectory to obtain an approximate solution of the landau - lifshitz - gilbert equation .
    			\end{tabularx}
    			\\
    			\hline
    			\begin{tabularx}{14.2cm}[c]{@{}X@{}}
			\textbf{\textit{GSS}}: 
				the finite difference or finite element computation of the demagnetizing fields and the time integration of the landau - lifshitz - gilbert equation requires a considerable computational effort .
			in this letter we propose a \hlc[cyan!30]{convolutional neural network} to reduce the dimensionality of thin film \hlc[yellow!40]{magnetization} and show how latent space dynamics can be applied to predict the magnetic response of magnetic thin film elements .
			in addition , we demonstrate that we can handle complex parameterizations in reduced spaces , and advance simulations \hlc[pink!80]{in time} by integrating in the latent space with a second network .
    			\end{tabularx}\\
    			\hline
    			\begin{tabularx}{14.2cm}[c]{@{}X@{}}
			\textbf{\textit{QA}}: 
				(1) What techniques are used to model the magnetization dynamics? [deep neural networks/ convolutional neural networks]; (2) 				The simulation speed become more quickly or slowly by using such technique? [quickly]
    			\end{tabularx}
				\\ \bottomrule      
			\end{tabular}
			\caption{Selected case study 			which shows how the citations help summarization.
			The same color denotes the same key concepts shared across documents.}
			\vspace{-4mm}
			\label{tab:case}
		\end{center}
	\end{table*}

\textbf{Ablation study.}
We also perform ablation studies to investigate the influence of different modules in our proposed models.
For unsupervised MUS, we remove the finetuning process on SciBERT, the position-aware weight, and, the sentence-level weight, the section-level weight, and the document-level in the multi-granularity graph during the importance calculation stage.
The third block in Table \ref{tab:comp_ROUGE_baselines} presents the results.
Specifically, RG-L score drops on both datasets after the sentence level, the section level, and the document level granularity are removed, indicating that multi-granularity information helps the model identify important sentences during the unsupervised phrase.

For supervised GSS, modules are tested in three ways: 
 (1) we first remove the graph encoder and directly use the average of sentence representations as the document representation;
(2) we remove the gated sentence encoder;
(3) we remove the graph information fusion module; and
(4) we remove the link prediction task so that the multi-task framework becomes a single task.
The results are shown in the final block of Table \ref{tab:comp_ROUGE_baselines}.
The performances of these models are worse than that of GSS in all metrics, demonstrating the effectiveness of GSS. 
Specifically, graph encoder improves GSS by 1.12 RG-L and 0.74 BERTSCORE on SSN (transductive), demonstrating that adjacent information in the citation graph helps summarization in the target paper.

\subsection{Further Discussion}

\begin{figure}[tb]
	\centering
	\includegraphics[width=0.7\linewidth]{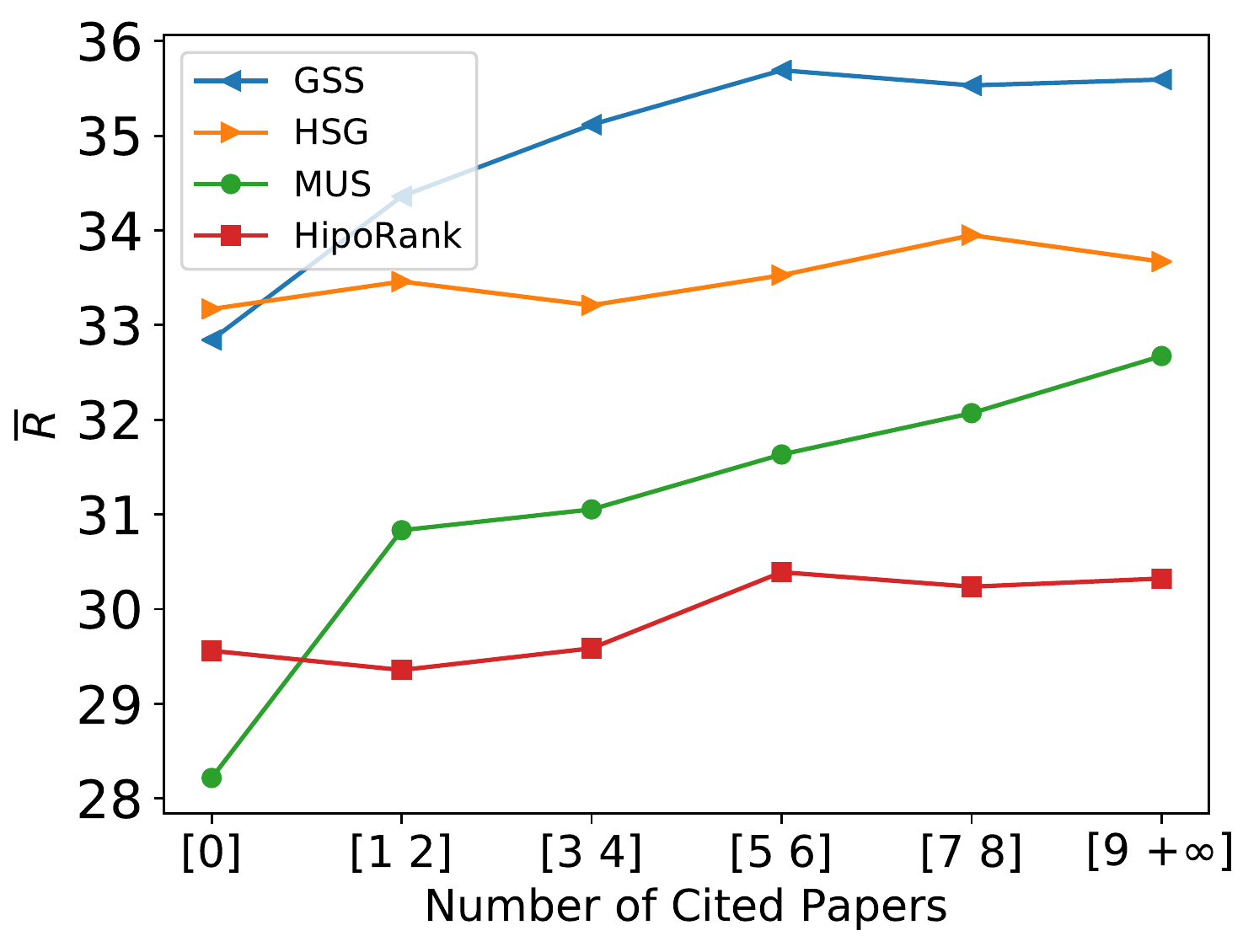}
	\caption{Relationships between the number of the cited papers in each ease (X-axis) and $\overline{R}$ (the average of RG-1, RG-2 and RG-L) of four models. 
	Best viewed in color.}
	\vspace{-2mm}
	\label{fig:citation}
\end{figure}

\textbf{Impact of the Number of Cited Papers.}
We further explore the impact of the number of cited papers in a target paper on the model performance.   
We divide SSN (inductive) test set into six groups according to the number of cited papers per target paper. 
As shown in Figure \ref{fig:citation}, the performance of \texttt{HipoRank} and \texttt{HSG} generally remains the same, demonstrating the stable performance regardless to the number of referred papers in summarization. 
Another interpretation is that they cannot benefit from the richer citation information available in groups with more cited papers.
On the contrary, as more papers are involved, MUS and GSS can generally make better summarizations. This indicates the capability of MUS and GSS in handling papers with rich citation graph information. They can extract the related information for making a more accurate summarization.

	\begin{table}[tb]
	\small
	\centering{\begin{tabular}{lcc}
			\toprule
			Model& SSN (inductive) &SSN (transductive)\\
			\midrule
			GSS & \textbf{81.67}$\pm$0.42 & \textbf{79.05}$\pm$0.34\\
			GSS w/o MT & 79.94$\pm$0.51 & 77.88$\pm$0.33 \\
			GSS w/o GSE & 80.66$\pm$0.45 & 78.22$\pm$0.53 \\
			GSS w/o GIF & 81.31$\pm$0.39 & 78.99$\pm$0.36 \\
			\bottomrule
	\end{tabular}}
	\caption{Link prediction performance accuracy (\%) of our model and ablation models.
	MT, GSE, GIF denotes Multi-task, Gated Sentence Encoder, Graph Information Fusion, respectively.}
	\label{tab:link}
\end{table}
	
\textbf{Performance of Link Prediction Task.}
We report the performance of the link prediction task as an auxiliary result for GSS, shown in Table~\ref{tab:link}.
GSS w/o multi-task is the baseline where we finetune the pre-trained SciBERT model on the training dataset and directly use it for link prediction.
The settings of GSS w/o gated sentence encoder and GSS w/o global transformer are the same as in the ablation study section.
All the experiments are repeated 10 times and we report the average accuracy with standard deviation.
It can be seen that GSS performs worst without the multi-task framework, demonstrating the effectiveness of the summarization task on the link prediction task.
Removing gated sentence encoder and global transformer also brings harm to GSS, proving it is necessary to use the document representation to polish the sentence representation.

\begin{table}[tb]
		\small
		\centering
		\begin{tabular}{l ccc}
			\toprule
			\multirow{3}{*}{Models} & \multicolumn{3}{c}{\textsf{ SSN (inductive)}}  \\
			\cmidrule(lr){2-4}
			& RG-1  & RG-2  & RG-L \\
		
			\midrule
		
			MUS (SciBERT) &  \textbf{43.89} & 13.07 & \textbf{39.04} \\
		
       MUS (BERT)  & 43.10 & 12.48 & 38.59 \\ 
        MUS (ROBERTa)  & 43.85 & \textbf{13.31} & 38.89 \\ 
        MUS (BART)  & 43.66 & 12.88 & 38.65 \\
        \midrule
        GSS (SciBERT) &  \textbf{47.71} & \textbf{16.78} & \textbf{42.04} \\
       GSS (BERT)  & 47.43 & 16.03 & 41.01 \\ 
        GSS (ROBERTa)  & 47.68 & 16.41 & 41.86 \\ 
        GSS (BART)  & 47.45 & 16.32 & 41.87  \\
       			\bottomrule
		\end{tabular}
		\caption{Performance of our models with different pretrained embeddings.}
		\label{tab:embedding}
	\end{table}

\textbf{Performance of Different Sentence Embeddings.}
Table \ref{tab:embedding} shows the results of our model with different embedding methods. 
MUS and GSS perform consistently across different embedding methods, demonstrating that the good performance of our model does not rely on specific embedding methods. 
Moreover, if we compare across different methods, sentence embeddings that are suitable for similarity computations such as SciBERT and RoBERTa do perform better than other choices.

\section{Conclusion}
Researchers usually write an abstract of a paper by referring to some examples, especially from a large number of papers on the same topic. 
Hence, in this paper, we propose to leverage citation graphs to enhance scientific paper extractive summarization.
Concretely, we propose a multi-granularity unsupervised summarization model and a graph-based supervised summarization model, both of which outperform the state-of-the-art baselines by a large margin.
In the future, we will explore the effect of other graph structure information on the summarization task such as pairwise distance information.


\section*{Acknowledgments}

We would like to thank the anonymous reviewers for their constructive comments. 
This work was supported by the SDAIA-KAUST Center of Excellence in Data Science and Artificial Intelligence (SDAIA-KAUST AI).
This publication is based upon work supported by the King Abdullah University of Science and Technology (KAUST) Office of Research Administration (ORA) under Award No FCC/1/1976-44-01, FCC/1/1976-45-01, URF/1/4663-01-01, and BAS/1/1635-01-01.

\bibliography{anthology,custom}
\bibliographystyle{acl_natbib}

\end{document}